\pdfoutput=1

\documentclass[11pt]{article}
\usepackage[preprint]{acl}

\usepackage{times}
\usepackage{latexsym}
\usepackage{booktabs}
\usepackage[misc]{ifsym}

\usepackage{amsmath}
\usepackage{amssymb}

\usepackage[T1]{fontenc}

\usepackage[utf8]{inputenc}

\usepackage{microtype}
\usepackage{graphicx}
\usepackage{subcaption}

\usepackage{inconsolata}

\title{Harder Tasks Need More Experts: Dynamic Routing in MoE Models }

\author{Quzhe Huang\textsuperscript{1}\thanks{\quad  Equal Contribution. \quad \Letter \   Songfang Huang and Yansong Feng are the corresponding authors.} , Zhenwei An\textsuperscript{2}$^*$, Nan Zhuang\textsuperscript{2}$^*$, Mingxu Tao\textsuperscript{1}, \\ \textbf{Chen Zhang\textsuperscript{1}},  \textbf{Yang Jin\textsuperscript{1}}, \textbf{Kun Xu\textsuperscript{3}}, \textbf{Kun Xu\textsuperscript{3}}, \textbf{Liwei Chen\textsuperscript{3}},  \\  \Letter \textbf{Songfang Huang\textsuperscript{2}}, \Letter \textbf{Yansong Feng\textsuperscript{1}} \\
  \textsuperscript{1}Peking University \quad \textsuperscript{2}AGIBang \quad \textsuperscript{3}Kuaishou Technology \\
  {\tt \{huangquzhe,anzhenwei\}@pku.edu.cn} \ {xmzhuang@agibang.ai} \\
  {\tt sfh@agibang.ai \quad fengyansong@pku.edu.cn} \\}

\begin{document}
\maketitle
\begin{abstract}

In this paper, we introduce a novel dynamic expert selection framework for Mixture of Experts (MoE) models, aiming to enhance computational efficiency and model performance by adjusting the number of activated experts based on input difficulty. Unlike traditional MoE approaches that rely on fixed Top-K routing, which activates a predetermined number of experts regardless of the input's complexity, our method dynamically selects experts based on the confidence level in expert selection for each input. This allows for a more efficient utilization of computational resources, activating more experts for complex tasks requiring advanced reasoning and fewer for simpler tasks. 
Through extensive evaluations, our dynamic routing method demonstrates substantial improvements over conventional Top-2 routing across various benchmarks, achieving an average improvement of 0.7\% with less than 90\% activated parameters. Further analysis shows our model dispatches more experts to tasks requiring complex reasoning skills, like BBH, confirming its ability to dynamically allocate computational resources in alignment with the input's complexity.
Our findings also highlight a variation in the number of experts needed across different layers of the transformer model, offering insights into the potential for designing heterogeneous MoE frameworks. The code and models are available at https://github.com/ZhenweiAn/Dynamic\_MoE.
\end{abstract}
\section{Introduction}
To effectively increase the model's parameter size, researchers have proposed the Mixture of Experts (MoE) framework~\cite{DBLP:conf/iclr/ShazeerMMDLHD17,DBLP:conf/iclr/LepikhinLXCFHKS21}. By setting up multiple experts to enhance the model's overall capacity, MoE models selectively activate a subset of parameters for use, thereby achieving more efficient parameter utilization. With the same number of activated parameters, MoE models substantially outperform dense models in performance, achieving exceptional results in tasks such as QA and machine translation~\cite{DBLP:journals/corr/abs-2109-10465}.

Most MoE frameworks adopt a routing mechanism that dispatches a fixed number of experts for every input~\cite{DBLP:journals/jmlr/FedusZS22,DBLP:conf/icml/DuHDTLXKZYFZFBZ22}. The most famous method is Top-K routing~\cite{DBLP:conf/iclr/ShazeerMMDLHD17}, which initially calculates the probability of each expert being suited to the current input and then activates the Top-K suitable experts. Empirically, previous works~\cite{DBLP:conf/iclr/LepikhinLXCFHKS21} activate two experts per token, as activating more experts offers limited improvements in model performance but substaintially increases training overhead. Most of the subsequent studies~\cite{zoph2022st,DBLP:conf/icml/LewisBDGZ21} can be seen as variants of Top-K routing, where different constraints are introduced to ensure that the number of tokens processed by different experts is as balanced as possible. Almost all these efforts activate a fixed number of experts.

The Top-K routing, though achieves good performance on downstream tasks, overlooks the different difficulties of inputs. Compared with simpler input, the more challenging input, e.g, tasks that require complex reasoning or logic inference,  might need more parameters to solve. Dispatching experts equally across inputs could lead to computational waste on simpler tasks and insufficient computational resources for more difficult ones.

\begin{figure*}[t]
\begin{subfigure}{0.5\textwidth}
    \includegraphics[width=\textwidth]{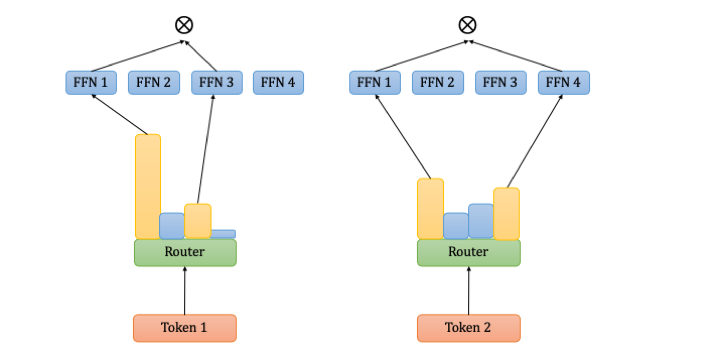}
    \caption{Top-K routing}
    \label{fig:first}
\end{subfigure}
\hfill
\begin{subfigure}{0.5\textwidth}
    \includegraphics[width=\textwidth]{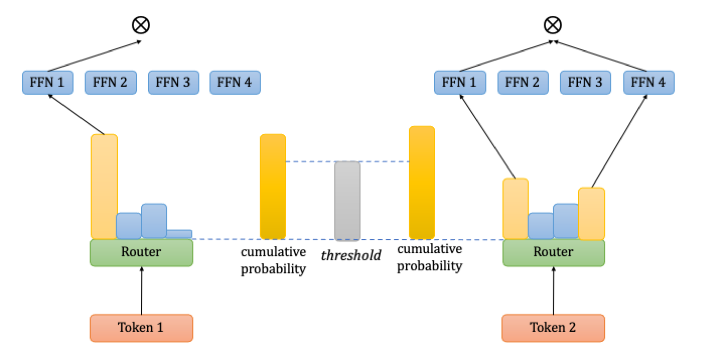}
    \caption{Top-P routing}
    \label{fig:second}
\end{subfigure}
\caption{Comparison between Top-K routing mechanism and Top-P routing mechanism. (a) Each token selects fixed K=2 experts with Top-K routing probabilities. (b) In Top-P routing mechanism, each token selects experts with higher routing probabilities until the cumulative probability exceeds threshold.}
\end{figure*}

To fully leverage the potential of MoE models, we propose a dynamic routing mechanism that adjusts the number of required experts based on the confidence level in the expert selection. When the model deems the currently selected experts as insufficient, it activates more experts. Specifically, we first compute a probability distribution for selecting experts. If the highest probability for an expert exceeds a predefined threshold $p$, indicating high confidence, we activate only that one expert. Otherwise, we progressively include additional experts until the cumulative probability of the selected experts exceeds the threshold $p$. This approach allows for a dynamic selection of experts, with the number of experts adjusted according to the input's complexity.

Evaluation across multiple common benchmarks has revealed that our method substantially outperforms MoE models based on Top-K routing. Compared with Top-2 routing,  our dynamic routing achieves an average improvement of 0.7\% with less than 90\% activated parameters. Further analysis has shown that our dynamic routing mechanism activates more experts in tasks requiring complex reasoning like BBH~\cite{DBLP:conf/acl/SuzgunSSGTCCLCZ23}, while using fewer experts in relatively easier tasks such as Hellaswag~\cite{DBLP:conf/acl/ZellersHBFC19}, confirming that our method indeed dynamically allocates experts based on the difficulty of the input. Token-level analysis indicates that tokens with ambiguous semantics are more challenging for the model, typically activating more experts. Another interesting finding is that the number of experts needed varies across different layers of the transformer. Lower layers require more experts for combination, while the top layer needs only one. This may relate to the \textit{over-thinking} phenomenon~\cite{kaya2019shallow} widely observed in deep neural networks.

Our contributions can be summarized as follows:
\begin{enumerate}
    \item  We proposed a dynamic routing strategy that can adjust the number of activated experts based on the input difficulty dynamically.
    \item We empirically validate that our proposed method is efficient in both training and inference, outperforming Top-2 routing while activating fewer experts.
    \item We observe that for MoE models, the number of experts needed to be activated varies across different layers. This finding could help design heterogeneous MoE frameworks.
\end{enumerate}

\section{Method}

In this section, we first briefly introduce the MoE model with Top-K routing strategy, which activates a fixed number of experts for each token. As Top-K routing ignores the varying difficulty of different inputs and the different requirements for experts at different layers, we propose a dynamic routing mechanism that adjusts the number of activated experts according to the complexity of inputs.
To avoid activating too many parameters through the dynamic routing mechanism, we also introduce a dynamic loss to encourage the model to activate only the necessary experts.

\subsection{Top-K Routing MoE}
In a Transformer model, the MoE layer is applied independently per token and replaces the feed-forward (FFN) sub-block of the transformer block~\cite{DBLP:conf/iclr/LepikhinLXCFHKS21}. For an MoE layer with $N$ experts, $E = \{e_1, e_2, ..,e_N\}$, an input $\mathbf{x}$ will be sent to the experts and the output of the MoE layer is the weighted average of the experts':

\begin{equation}
    MoE(\mathbf{x}) = \sum_{i=1}^N g_i(\mathbf{x}) * e_i(\mathbf{x})
\end{equation}
where $g_*(\mathbf{x})$ is computed by a routing network that determines the contribution of each expert to the final output. In consideration of computing efficiency, a token is dispatched to limited experts. Thus for most experts, the corresponding $g_*(\mathbf{x})$ is zero meaning that the token is not dispatched to that expert.

To obtain $g_*(\mathbf{x})$, we first compute the probability $\mathbf{P}$ of selecting each expert for input $\mathbf{x}$:
\begin{equation}
    \mathbf{P} = Softmax(\mathbf{W_r} \cdot \mathbf{x}^T)
    \label{eq:P}
\end{equation}
where $\mathbf{W_r} \in {N \times d}$ is a learnable parameter and $d$ is the dimension of the input $\mathbf{x}$. $\mathbf{P}$ is a vector of size $N$ and $P_i$ represents the probability of selecting the $i^{th}$ expert $e_i$ to calculate the input $\mathbf{x}$.

Top-K routing selects the k experts, whose probabilities are the highest k in $\mathbf{P}$. Then the probabilities of the selected experts are normalized and the weights of the remaining experts are set to zero, indicating they are not activated. The corresponding calculation of $g_*(\mathbf{x})$ is as follows:
\begin{equation}
	g_i(\mathbf{x}) = 
        \begin{cases}
	\frac{P_i}{\sum_{j \in TopK(\mathbf{P})} P_j}, &\ i \in TopK(\mathbf{P})\\
	0, &\ i \notin TopK(\mathbf{P})
        \end{cases}%
\end{equation}
where $TopK(\mathbf{P})$ returns the indices of the largest k elements in $\mathbf{P}$. 

Top-K routing is initially proposed by \cite{DBLP:conf/iclr/ShazeerMMDLHD17}, and subsequently, numerous studies have built upon it with improvements. The following works~\cite{DBLP:conf/iclr/LepikhinLXCFHKS21, DBLP:conf/iclr/Zuo00KHZGZ22} introduce constraints aimed at ensuring a more balanced workload among the experts during training. The core of these works remains to select the most suitable experts for each token under specific constraints, based on the probability distribution $\mathbf{P}$ calculated in Equation~\ref{eq:P}. And the number of experts dispatched for each token is fixed across all these studies. Empirically, the value of k is set to 2, serving as a trade-off between training costs and model capabilities.

\subsection{Dynamic Routing MoE}

Although the Top-K routing strategy has shown promising performance, its assumption that an equal number of experts should be dispatched for each token overlooks the variability in difficulty across different inputs. Moreover, since a fixed number of experts are activated at every layer of the transformer, this approach neglects the differences in representations across layers, potentially requiring a different number of experts for different layers. 

To address these issues and make use of model parameters more efficiently, we propose a dynamic routing strategy based on model confidence. Unlike the Top-K routing, which selects a fixed number of experts, our method allows the model to assess whether the currently selected experts are sufficient. If not, it continues to incorporate more experts. 

Specifically, we regard that $\mathbf{P}$ in Equation~\ref{eq:P} reflects the confidence level of input $\mathbf{x}$ in selecting different experts. In other words, $P_i$ represents how confident the model is that the $i^{th}$ expert can adequately handle input $\mathbf{x}$. If the highest probability in $\mathbf{P}$ is sufficiently large, then we may only need to use the corresponding expert. However, if the highest probability is not large enough, we need to add more experts to increase the reliability of processing $\mathbf{x}$. We keep adding experts until the sum of the probabilities of the selected experts exceeds a specific threshold $p$, at which point we consider the model confident enough that these experts can effectively process the input $\mathbf{x}$. We add new experts in descending order of their probabilities in $\mathbf{P}$ to minimize the number of activated experts as much as possible. 

Formally, we first sort the elements in $\mathbf{P}$ from highest to lowest, resulting in a sorted index list $I$. Then we find the smallest set  of experts whose cumulative probability exceeds the threshold $p$, and the number of selected experts $t$ is calculated by:
\begin{equation}
    t = \mathop{arg min}_{k \in \{1..., N\}} \ \ \sum_{j <= k}{P_{i,j}}  \geq p
    \label{eq:t}
\end{equation}
where $p$ is the threshold that controls how confident the model should be when stopping adding more experts. $p$ is a hyper-parameter whose range is from 0 to 1. The higher the $p$ is, the more experts will be activated.

In dynamic routing mechanism, the calculation of $g_*(\mathbf{x})$ is:
\begin{equation}
	g_i(\mathbf{x}) = 
        \begin{cases}
	P_i &\ e_i \in S \\
	0, &\ e_i \notin S
        \end{cases}%
\end{equation}
where $S$ is the set of selected experts controlled by $t$ in Equation~\ref{eq:t}:
\begin{equation}
    S = \{e_{I_1}, e_{I_2} ... e_{I_t}\}
    \label{eq:S}
\end{equation}

\subsection{Loss}

\paragraph{Dynamic Loss}

There is a risk associated with our dynamic routing mechanism: it could assign low confidence to all experts, thereby activating a larger number of experts to achieve better performance. Suppose $\mathbf{P}$ is a uniform distribution and we set the hyper-parameter $p$ to 0.5, then the model would activate up to half of the experts. This goes against the original intention of the MoE framework, which is to scale the model with great efficiency.

To prevent dynamic routing from using too many parameters to cheat and losing its ability to selectively choose experts, we introduce a constraint on $\mathbf{P}$. We expect the routing mechanism to select a small set of necessary experts, therefore, we aim to minimize the entropy of the distribution $\mathbf{P}$, ensuring that every token can focus on as less specific experts as possible. Our dynamic loss is designed to encourage the routing mechanism to select the minimal necessary set of experts, which is formalized as:

\begin{equation}
    Loss_{d} = - \sum_{i=1}^N {P_i * log (P_i)}
\end{equation}

\paragraph{Load Balance Loss}
MoE models typically require distributed training, where different experts are deployed across various nodes. To avoid scenarios where some nodes are fully utilized while others are underutilized, thereby impacting training efficiency, it is generally desirable for the number of tokens processed by different experts to be roughly the same. Furthermore, previous study~\cite{DBLP:conf/iclr/Zuo00KHZGZ22} has shown that evenly activated experts in an MoE layer can lead to better performance. To achieve balanced loading among different experts, we have also incorporated a load-balance loss, \(Loss_b\), which is widely used in previous works~\cite{DBLP:conf/iclr/LepikhinLXCFHKS21, DBLP:journals/jmlr/FedusZS22}

\begin{equation}
    Loss_b = N * \sum_{i=1}^N{f_i * Q_i}
\end{equation}
where $f_i$ is the fraction of the tokens choosing expert $e_i$  and $Q_i$ is the fraction of the router probability allocated for expert $e_i$. For a sequence containing $M$ tokens, $f_i$ and $Q_i$ are calculated as:
\begin{equation}
    f_i = \frac{1}{M}\sum_{j=1}^M{1 \{e_i \in S^j\}}
\end{equation}
\begin{equation}
    Q_i = \frac{1}{M}\sum_{j=1}^n{P_i^j}
\end{equation} 
where $S^j$ is the set of activated experts for token $j$, which is calculated by Equation~\ref{eq:S}, and $P^j$ is the probability of selecting each experts for token $j$, calculated by Equation~\ref{eq:P}.

\paragraph{Final Loss}
Our model is a generative model that uses next token generation as the training objective. We denote this loss as $Loss_{lm}$. Our final loss is a combination of the language model loss, dynamic loss, and load-balance loss:

\begin{equation}
    Loss = Loss_{lm} + \alpha Loss_b + \beta Loss_d
\end{equation}
where $\alpha$ and $\beta$ are hyper-parameters to adjust the contribution of the load balance loss and dynamic loss, respectively. In our experiment, we set $\alpha$ as 1e-2 and $\beta$ is set as 1e-4.

\section{Experiments}

\subsection{Settings}
\subsubsection{Training data}
We use RedPajama\cite{together2023redpajama} as our training data, which is a fully open-source implementation of the LLaMa dataset. RedPajama data consists of diverse sources including the common crawl (CC), C4, github, Wikipedia, books, arxiv and Stackexchange.  In our main experiments, we train all models for 100B tokens. 

\subsubsection{Model Settings}
The model architecture follows LLaMA\cite{DBLP:llama}. We use Llama2 tokenizer whose vocabulary size is 32,000. The number of transformer layers is 24 and the hidden dimension is 1024. Each MoE layer has 16 experts. Under this configuration, dense model has approximately 374M parameters. Each MoE model has 3.5B total parameters. Only 374M parameters are activated in MoE-Top1 and 581M parameters are activated in MoE-Top2. More detailed model and training settings are shown in Appendix~\ref{sec:appendix}.

\begin{table*}[]
\small
\centering
\begin{tabular}{@{}cccccc@{}}
\toprule
               & Dense(374M) & Dense(570M)   & MoE-Top1 & MoE-Top2      & MoE-Dynamic   \\ \midrule
PIQA~\cite{DBLP:conf/aaai/BiskZLGC20}           & 64.3        & 65.9          & 67.3     & \textbf{68.1} & \textbf{68.1}          \\
Hellaswag~\cite{DBLP:conf/acl/ZellersHBFC19}      & 36.1        & 39.6          & 42.3     & 43.9          & \textbf{44.3} \\
ARC-e~\cite{DBLP:journals/corr/abs-2102-03315}          & 37.9        & 37.6          & 39.5     & \textbf{40.4} & 39.9 \\
Commonsense QA~\cite{DBLP:conf/naacl/TalmorHLB19} & 32.2        & 31.7          & 30.3     & 32.1          & \textbf{33.6} \\
BBH~\cite{DBLP:conf/acl/SuzgunSSGTCCLCZ23}            & 22.3        & 22.1          & 23.0     & 23.3          & \textbf{25.6} \\ \midrule
Avg            & 38.6        & 39.4          & 40.5     & 41.6          & \textbf{42.3} \\ \bottomrule
\end{tabular}
\caption{Performance on downstream tasks. The best result for each task is emphasized in \textbf{bold}. }
\label{tab:main_result}

\end{table*}

\subsubsection{Evaluation}
We use opencompass\footnote{https://github.com/open-compass/OpenCompass/} to evaluate our model.

\subsubsection{Experiment Models}
We train several variants of our architecture from scratch using the above model settings.
\paragraph{Dense} We use dense models as our baseline. In dense models, each transformer layer is composed of a multi-head attention layer and a standard Feed Forward Network. We implement two Dense models: Dense(374M) and Dense(570M) by setting the hidden dimensions to 1024 and 1280, respectively.

\paragraph{MoE-Top1 / Top2} The MoE models with Top-K routing, where K = 1 and 2, respectively. Only language modeling loss, $Loss_{lm}$, and load-balance loss $Loss_{b}$ are used for training.  The MoE-Top1 could be seen as a re-implementation of Switch Transformer~\cite{DBLP:journals/jmlr/FedusZS22} and the MoE-Top2 is a re-implementation of Gshard\cite{DBLP:conf/iclr/LepikhinLXCFHKS21}. The activated parameters of MoE-Top1 and MoE-Top2 are nearly the same as Dense(374M) and Dense(570M), respectively.

\paragraph{MoE-Dynamic}
MoE-Dynamic model uses our dynamic adaptive routing mechanism, activating a various number of experts depending on the input token representation. The threshold $p$ in our routing mechanism is 0.4. During inference, MoE-Dynamic model activates no more than 2 experts, which means it uses fewer parameters than MoE-Top2.

\subsection{Main Results}

Table~\ref{tab:main_result} shows the performance of different models on downstream tasks. Overall, the MoE models outperform the Dense models. Among all the MoE variants, our proposed Dynamic Adaptive MoE demonstrates the best performance, achieving at least a 0.7\% higher score on average compared to other models.

We first compare models with an equal number of activated parameters. It is observed that MoE-Top1 outperforms the Dense model with 374M parameters by an average of 1.9\% score, and MoE-Top2 surpasses the Dense model with 570M parameters by 2.2\% score. This indicates that, with the same number of activated parameters, MoE models substantially outshine their corresponding Dense counterparts.

When comparing models with the same architecture, we generally observe a positive correlation between model performance and the number of activated parameters. For Dense models, the model with 570M parameters outperforms the model with 374M parameters by 0.8\% score on average. 
Similarly, among models using the MoE architecture with a fixed number of activated experts, MoE-Top2, which activates two experts, reaches an average of 41.6\% score, outperforming MoE-Top1, which only activates one expert, by 1.1\% score. In fact, MoE-Top2 performs better than MoE-Top1 in all subtasks, demonstrating the rule of more parameters leading to better performance.

However, our proposed Dynamic Routing mechanism breaks this rule. As shown in Table~\ref{tab:activated_experts_in_test}, the average number of activated experts in the MoE-Dynamic during evaluation phases is less than two, meaning it activates fewer parameters than MoE-Top2. Yet, as shown in Table~\ref{tab:main_result}, compared to MoE-Top2, MoE-Dynamic achieves comparable or even better performance on nearly all the tasks and outperforms MoE-Top2 by 0.7\% score on average.
MoE-Dynamic obtains better performance, indicating that our dynamic routing mechanism can allocate the necessary experts for different inputs more reasonably and make use of parameters more efficiently.

\begin{figure}[t]
\centering
\includegraphics[scale=0.45]{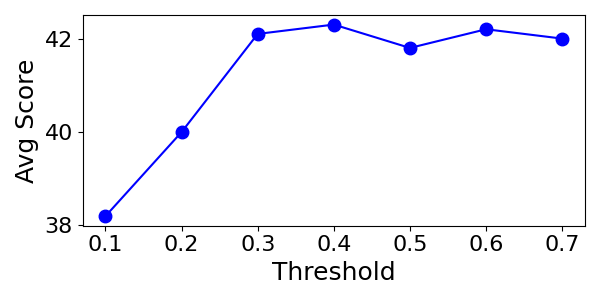}
\caption{Average scores of MoE-Dynamic with different threshold $p$ on downstream tasks}
\label{fig:p_influence_performance}
\end{figure}
\subsection{Effect of Threshold $p$}
Threshold \(p\) is a hyper-parameter used to control the dynamic routing mechanism. Training models from scratch with different values of \(p\) is resource-intensive. Hence, we explore the impact of this hyper-parameter by performing inference on a pre-trained model with varying values of \(p\) from 0.1 to 0.7. Table~\ref{fig:p_influence_performance} demonstrates the average performance on downstream tasks with different $p$.

The table reveals that when \(p\) is too low, like 0.1 and 0.2, the model's performance on downstream tasks markedly decreases due to the activation of too few experts. Conversely, once \(p\) surpasses a certain threshold, the model's performance stabilizes, and the impact of this parameter on downstream tasks will become minimal.

\section{Efficiency of Dynamic Routing}

The greatest advantage of MoE models is their ability to efficiently scale to larger models. The Top-K routing mechanism controls the number of parameters used by the entire model by activating a fixed number of experts. In contrast, our proposed dynamic routing mechanism removes the limitation of a fixed number of experts. Naturally, there may be concerns that our method might assign too many experts to each token. To address these concerns, we demonstrate the efficiency of the dynamic routing mechanism from both training and inference perspectives.

\subsection{Efficient Training}
\begin{figure}[t]
\centering
\includegraphics[scale=0.5]{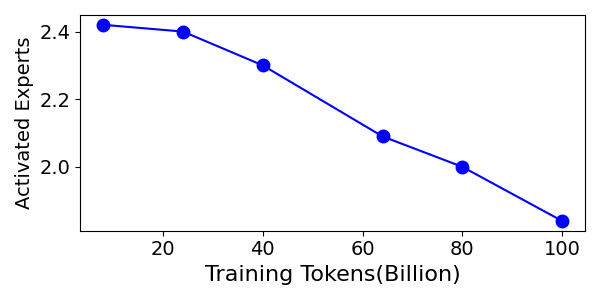}
\caption{Average activated experts number across training procedure.}
\label{fig:training_process_experts}
\end{figure}

We sample 1000 pieces of data from different sources within Redpajama and calculate the average number of experts activated per token at different stages of training.  Figure~\ref{fig:training_process_experts} shows the change in the average number of experts activated throughout the training process of 100B tokens. From the figure, we can see that the number of experts activated per token decreases over time. In the early stages of training, dynamic routing assigns more experts to each token, but after 60B tokens, the average number of activated experts is already less than 2. Table~\ref{tab:activated_experts_in_train} displays the number of experts activated by MoE-Dynamic at the end of the 100B training. It is evident that across all data sources, the number of experts activated by MoE-Dynamic is less than 2.

Recently, the amount of tokens used in training for large language models far exceeds 100B, for instance, Pythia uses 300B tokens, and Llama2 uses 2T tokens. If we continue to train on an even larger scale corpus, the average number of parameters used throughout the training process is guaranteed to be lower than that of Top2-Routing.

\begin{table}[]
\small
\centering
\begin{tabular}{@{}lcc@{}}
\toprule
Sources       & Ratio & Activated Experts \\ \midrule
CC            & 67\%  & 1.82              \\
C4            & 15\%  & 1.84              \\
Github        & 4.5\% & 1.88              \\
Wiki          & 4.5\% & 1.78              \\
Book          & 4.5\% & 1.73              \\
Arxiv         & 2.5\% & 1.90              \\
StackExchange & 2\%   & 1.79              \\ \midrule
Avg           & 100\% & 1.82              \\ \bottomrule
\end{tabular}
\caption{Average activated experts in different parts of the training corpus. }
\label{tab:activated_experts_in_train}

\end{table}

\subsection{Efficient Inference}

To further explore whether our proposed method is efficient in inference, we calculate the average number of experts activated by the model across different downstream tasks. For every question, we use the template from the evaluation to concatenate the question with the gold answer into a complete input and truncate the tokens exceeding 2048 to fit our model's maximum input length. Table~\ref{tab:activated_experts_in_test} shows the average number of experts activated per token across various downstream tasks. The result is averaged across all the layers of transformers and it is evaluated using the checkpoint trained on 100B tokens.

From the table, we can observe that across all five downstream tasks, the number of activated experts is less than two, averaging 1.76 activated experts, which is fewer than the fixed activation of two experts by the Top2 routing method. During the training phase, our method and Top2 routing are comparable in efficiency, but upon completion of training, our inference efficiency substantially outperforms Top2 routing. Given that models are mostly trained once with a greater burden placed on the subsequent deployment nowadays, the advantages of our method over traditional MoE routing mechanisms like Top2 become even more apparent.

\begin{table}[!t]
\centering
\small
\begin{tabular}{@{}lc@{}}
\toprule
Sources         & Activated Experts \\ \midrule
PIQA            & 1.72              \\
Winogrande      & 1.76              \\
ARC-e           & 1.73              \\
Commonsense QA & 1.74              \\
BBH             & 1.87              \\ \midrule
Avg             & 1.76              \\ \bottomrule
\end{tabular}
\caption{Average activated experts in different downstream tasks.}
\label{tab:activated_experts_in_test}

\end{table}

\section{What is Challenging Input?}

The motivation for designing dynamic routing is to enable the model to dynamically adjust the number of allocated experts based on the difficulty of the input. In this section, we will explore what kinds of inputs are considered challenging for the model from various perspectives.

\subsection{Tasks Requiring Reasoning}
From Table~\ref{tab:activated_experts_in_test}, we could observe that solving the BBH task requires activating an average of 1.87 experts,  more than the number needed for other tasks. BBH, which stands for BIG-Bench Hard, is a suite of 23 challenging BIG-Bench tasks. These tasks demand capabilities such as multi-hop reasoning, causal inference, logical deduction, and so on, making them substantially more difficult than normal NLP tasks~\cite{suzgun2022challenging}. 
Our model's use of more experts on BBH tasks implies that our method indeed can dynamically monitor task difficulty and apply more parameters to tackle more challenging tasks. Interestingly, as shown in Table~\ref{tab:main_result}, MoE-Dynamic, compared to MoE-Top2, sees the most improvement on BBH tasks. While the average improvement across all tasks is less than 1.0\%, the improvement on BBH is more than 2.0\%, which is more than double that of other tasks. This further illustrates that dynamically adjusting the number of activated experts is beneficial for solving downstream tasks, especially more challenging ones.

\subsection{Tokens with Ambiguous Semantics}

To further analyze what types of tokens are considered more challenging for a model, we examine the average number of experts activated for each token in the vocabulary across different contexts. 

We sample 1 million tokens from each part of the training dataset Redpajama, like arxiv and CC, resulting in a new corpus of a total of 7 million tokens. In this corpus, we calculate the average number of experts activated for each token in the vocabulary. To minimize the effect of randomness, we only consider tokens that appear more than 10,000 times in the corpus.

\begin{table}[!t]
\small
\centering
\begin{tabular}{@{}llc@{}}
\toprule
              & Examples & C-Words Ratio  \\ \midrule
Most Experts  &     tr, eq, mu, frac    &       10              \\
Least Expers &    to, that, and, show      &       51             \\ \bottomrule
\end{tabular}
\caption{The first column shows examples of tokens requiring the most experts and least experts. The last column shows the complete words ratio in these two groups of tokens.}
\label{tab:sub&complete-words}
\end{table}

Table \ref{tab:sub&complete-words} shows the number of complete words among the top 100 and bottom 100 tokens by the average number of experts activated, along with some examples.

Upon manually reviewing the 100 tokens that activate the most experts and the 100 tokens that activate the least, we observe an interesting phenomenon: Tokens with relatively definite semantics are considered easier by the model, activating fewer experts. In contrast, tokens with uncertain semantics are deemed more challenging and require more experts for processing.

Specifically, since our model's tokenizer is trained with Byte Pair Encoding (BPE), many tokens are not complete words but subwords. These subwords have vaguer semantics compared to full words because they can combine with many other subwords to form words with different meanings. For example, the subword 'tr' can lead to the formation of hundreds of words with varied meanings, such as tree, triple, train, trick, trouble, and so on. Due to the multitude of possible semantics, different meanings may require different experts for processing, making such subwords require a comprehensive understanding by more experts.

\section{Bottom Layers Need More Experts}

An intriguing observation from our study is that our model achieves superior performance while activating fewer parameters. As shown in Table~\ref{tab:activated_experts_in_test}, on all the tasks, our MoE-Dynamic activates an average of fewer than two experts. But it outperforms the MoE-Top2 in downstream tasks as shown in Table~\ref{tab:main_result}. This result is quite surprising, as performance on downstream tasks is typically correlated with the quantity of activated parameters. We attribute this unexpected phenomenon to our method's more proper allocation of the experts to be activated across different layers, employing more experts at lower levels and fewer at the top. This layer-wise dynamic allocation, as opposed to the fixed number of experts per layer, somewhat mitigates the common issue of overthinking in deep neural networks, thereby enhancing performance.

The overthinking refers to the situations where simpler representations of an input sample at an earlier layer, relative to the complex representations at the final layer, are adequate to make a correct prediction~\cite{kaya2019shallow}.
Previous works~\cite{liu2020fastbert,schwartz2020right,xin2021berxit} have demonstrated that shallower representations can achieve comparable, if not better, performance across various tasks than deeper representations. This could be due to deeper representations overfitting specific distributions, lacking generalizability, and being more vulnerable to attacks~\cite{hu2019triple,zhou2020bert}. It suggests that in some cases, acquiring a better shallow representation is more valuable than obtaining a more complex deep representation, which correlates to previous findings that removing top layers has a limited impact on the downstream tasks~\cite{sajjad2023effect}.

\begin{figure}[t]
\centering
\includegraphics[scale=0.5]{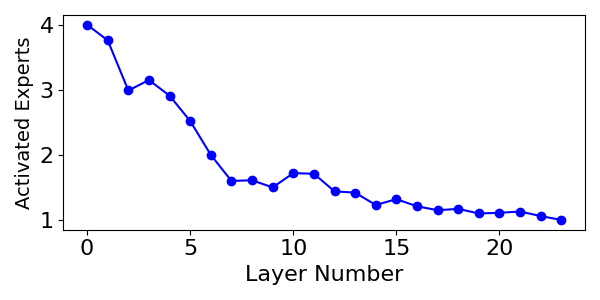}
\caption{Activated experts in different layers}
\label{fig:experts_per_layer}
\end{figure}

Compared with Top2 routing, our dynamic adaptive routing activates more experts at the bottom layers to obtain better shallow representations and use the simpler network in the top layers to alleviate the overthinking issue. Figure~\ref{fig:experts_per_layer} displays the number of experts activated per token at different layers\footnote{The results are evaluated using a checkpoint trained on 100B tokens.}.
From the figure, we observe a gradual decrease in the average number of experts activated per token with increasing layer depth. The lowest layer activates the most experts, up to 4 experts per token, enabling better shallow representations through a wider network, which is beneficial for various downstream tasks. At the topmost layer, the number of activated experts per token is reduced to even one. This phenomenon can avoid model being too complex and preserve generality in the final representation.

\section{Related Work}
The Mixture of Experts (MoE) model is initially introduced by~\cite{DBLP:journals/neco/JacobsJNH91}. 
Recent studies have demonstrated sparsely gated MoE models have substantial improvements in model capacity and efficiency, enabling superiors performance than dense models\cite{DBLP:conf/iclr/ShazeerMMDLHD17}. Particularly MoE has shown great potential with the integration of transformer architectures \cite{zoph2022st}.  \par
In previous MoE architectures, a static number of experts are activated regardless of the varying complexity presented by input tokens. Most of MoE models activate Top-1 or Top-2 experts \cite{DBLP:conf/iclr/LepikhinLXCFHKS21, DBLP:journals/jmlr/FedusZS22}, which could potentially limit the efficacy of MoE models. \par
There are works allocating various number of experts for input tokens. Expert-Choice MoE model selects Top-K tokens for each expert\cite{DBLP:conf/nips/ZhouLLDHZDCLL22}. However, in Expert-choice MoE model, the floating-point operations per second(FLOPS) in each MoE layer are the same. Previous work indicates that different MoE layers may need different FLOPS to achieve optimal performance\cite{DBLP:conf/acl/JawaharMLKALABG23}. \par

Different from these prior works, our dynamic routing mechanism can allocate more experts for complex tokens and fewer for simpler ones. Additionally, it strategically selects more experts in the lower layers and fewer in the upper layers, thereby minimizing computational redundancy. Experimental results demonstrate that this dynamic routing approach contributes to improvements in both the efficiency and performance of MoE models.

\section{Conclusion}

Our paper introduces a dynamic expert selection framework for Mixture of Experts (MoE) models, surpassing traditional fixed Top-K routing by adjusting expert activation based on input complexity. Our approach not only improves computational efficiency but also model performance, evidenced by obvious gains over conventional Top-K routing in our evaluations. Our findings reveal the framework's effectiveness at dynamically dispatching different numbers of experts, particularly for complex reasoning tasks, and suggest the potential for developing more challenging heterogeneous MoE models. In support of further research, we will open-source our models, contributing to advancements in the MoE domain.

\section*{Limitation}
Due to resource constraints, the size of the model we trained is limited, with only about 600M activation parameters, and the entire MoE (Mixture of Experts) model being just over 3B in size. However, \cite{dai2024deepseekmoe} has validated that within the MoE framework, conclusions drawn from smaller models can be generalized to larger models with more parameters. Hence, we believe our proposed dynamic routing method could also be effective in larger-scale models. Additionally, we have only trained on 100B tokens, which may not be sufficient for model training. Yet, given the same scale of training data, our method demonstrated superior performance, which also underscores the efficiency of our training process.
\bibliography{acl_latex}

\section{Detailed Training Setting}
\label{sec:appendix}
\subsection{Model Setting}
The model architecture follows LLaMA\cite{DBLP:llama}. We use Llama2 tokenizer whose vocabulary size is 32000. Unless specifically stated otherwise, we set the number of transformer layers to 24, the hidden dimension to 1024. We employ the multi-head attention mechanism with a total of 16 attention heads, where each head has a dimension of 64. We use SwiGLU\cite{DBLP:journals/corr/abs-2002-05202} in FFN layers. For initialization, all learnable parameters are randomly initialized with a standard deviation of 0.006. Each MoE layer has 16 experts, which have the same initialized parameters as a standard FFN. Under this configuration, each dense model has has approximately 374M parameters. Each MoE model has 3.5B total parameters. Only 374 parameters are activated in MoE-Top1 and 581M parameters are activated in MoE-Top2.
\subsection{Training Setting}
We adopt AdamW optimizer with first-moment decay $\beta_1 = 0.9$ and second-moment decay $\beta_2 = 0.95$. The weight decay is 0.1. The learning rate warms up from 0 to 3e-4 in the first 2000 steps and decays in the remaining steps using the cosine decay schedule to 3e-5. We set the context length to 2048 and adopt the batch size of 2048. 

\end{document}